\documentclass{article}
\usepackage[utf8]{inputenc}

\usepackage{amsmath}
\usepackage{bm}
\usepackage{algorithm}
\usepackage[noend]{algpseudocode}
\usepackage{amsthm}
\usepackage{amssymb}
\usepackage{graphicx}
\usepackage{subfigure}
\usepackage{soul}
\usepackage[square,sort,comma,numbers]{natbib}
\usepackage{float}
\usepackage{mathtools}

\makeatletter
\def\BState{\State\hskip-\ALG@thistlm}
\makeatother

\newtheorem{definition}{Definition}

\usepackage[final]{nips_2017}

\usepackage[utf8]{inputenc} 
\usepackage[T1]{fontenc}    
\usepackage{hyperref}       
\usepackage{url}            
\usepackage{booktabs}       
\usepackage{amsfonts}       
\usepackage{nicefrac}       
\usepackage{microtype}      
\usepackage{xcolor}

\title{NeurT-FDR: Controlling FDR by Incorporating   Feature Hierarchy }

\author{
Lin Qiu

\\

  Penn State University\\
  \texttt{\{luq7\}@psu.edu} \\

\And
Nils Murrugarra-Llerena

\\

    Snap Inc.\\
    \texttt{\{nmurrugarraller\}@snap.com} \\
  
\And
Vítor Silva

\\

    Snap Inc.\\
    \texttt{\{vsilvasousa\}@snap.com} \\

\And
Lin Lin

\\

    Penn State University\\
    \texttt{\{llin\}@psu.edu} \\

\And
Vernon M. Chinchilli

\\

    Penn State University\\
    \texttt{\{vchinchi\}@psu.edu} \\
}

\begin{document}
\maketitle

\begin{abstract}
Controlling false discovery rate (FDR) while leveraging the side information of multiple hypothesis testing is an emerging research topic in modern data science. Existing methods rely on the test-level covariates while ignoring possible hierarchy among the 
covariates. This strategy may not be optimal for complex large-scale problems, where hierarchical information often
exists among those test-level covariates. We propose \texttt{NeurT-FDR} which boosts statistical power and controls FDR for multiple hypothesis testing while leveraging the hierarchy among test-level covariates. Our method parametrizes the test-level covariates as a neural network and adjusts the feature hierarchy through a regression framework, which enables flexible handling of high-dimensional features as well as efficient end-to-end optimization. We show that \texttt{NeurT-FDR} has strong FDR guarantees and makes substantially more discoveries in synthetic and real datasets compared to competitive baselines.
\end{abstract}

\section{Introduction}

In modern statistics, from genetics, neuroimaging, to online advertising and finance, researchers routinely test thousands or millions of hypotheses at a time. Multiple Hypothesis Testing (MHT) considers those set of hypotheses together and perform statistical inferences simultaneously.  
The objective of MHT is to maximize the number of discoveries while controlling the FDR. In general, there are two statistical approaches to address multiple testing issue: (1) family-wise error rate approach (such as the Bonferroni correction), which controls the probability of making at least one false discoveries (Type I error), and (2) False Discovery Rate (FDR) approach, which controls the expected proportion of false discoveries. FDR is often more appealing for high-throughput data analysis, as it can substantially increase power while raising only a small fraction of false discoveries. The Benjamini and Hochberg linear step-up procedure~\cite{benjamini95}, and Storey's q value~\cite{storey04} are classical FDR-controlling methods.
However, those methods decide on the rejection rule based solely on the p values, and assume all hypotheses are exchangeable and each null hypothesis is equally likely to be true or false. 

\begin{figure}[H] 
\label{fig:illustration}
\centering
\includegraphics[width=1.0\textwidth]{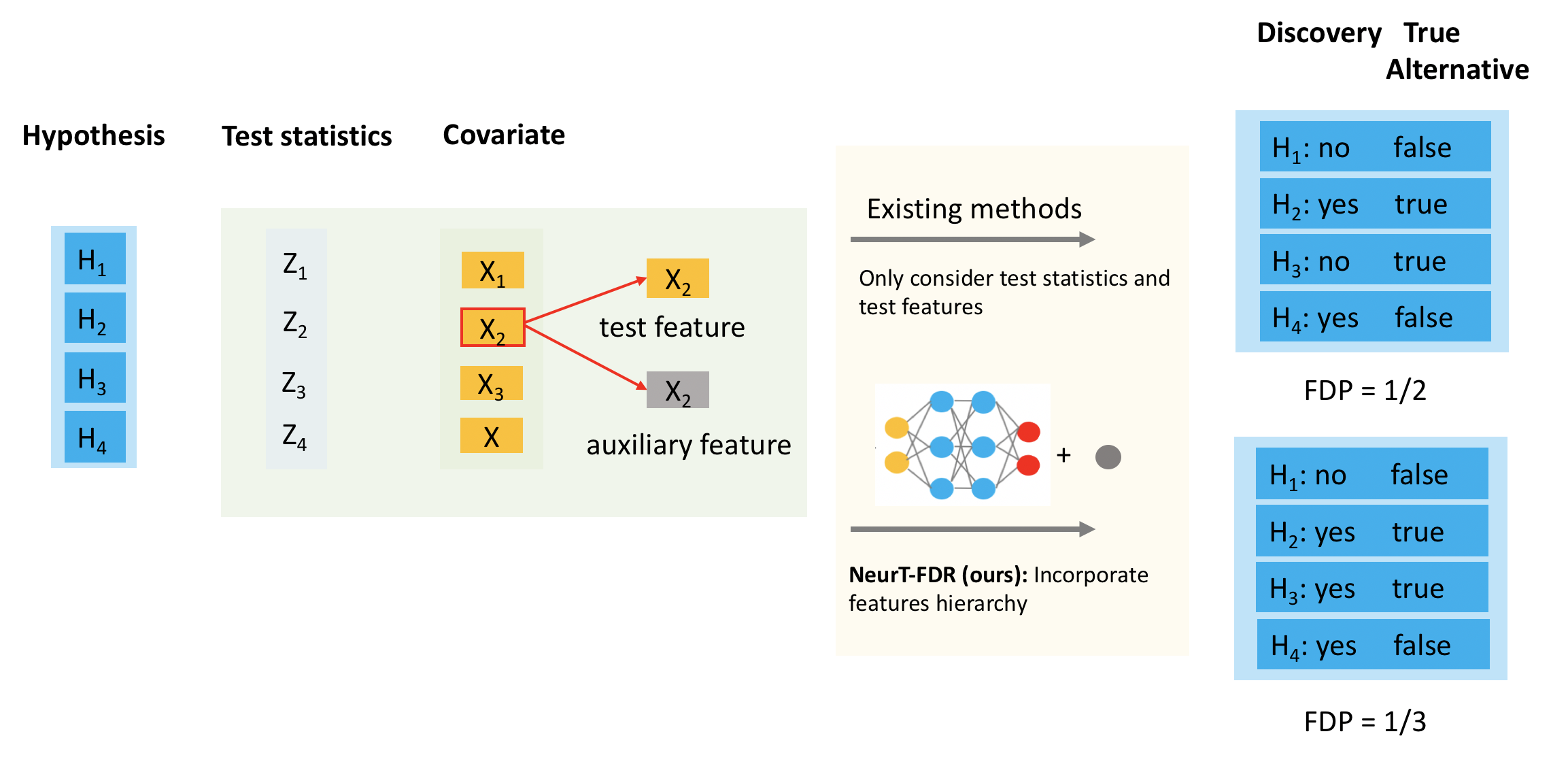}
\caption{For MHT problem, each hypothesis produces a test statistics. In addition, each hypothesis also has test-level covariates and the associated auxiliary features. Existing covariate-adaptive FDR methods considers only test-level covariates, while we propose a new method (NeurT-FDR) to incorporate both text-level covariates and auxiliary features.}
\label{fig:Xa}
\end{figure}

In many cases, we often have additional information or test-level covariates, which could be informative about the hypothesis tests. Such additional information typically contain certain auxiliary features. For example, gene length is such auxiliary feature for each gene in RNA-Seq differential expression analysis; another example is social media data, the user engagement metrics (such as the number of shares, the number of views, how much time spent on one post, etc.) are the auxiliary features for each post. Typically, such auxiliary features 
are of lower dimension than those text-level covariates. Recently, there has been a surge in developing covariate-adaptive FDR procedures that aim to improve the detection power while maintaining the target FDR~\cite{xia17,tansey18,zhang19}. However, those methods ignore the auxiliary features among the test-level covariates. In many cases and also in Section~\ref{sec:emp}, as illustrated in Figure~\ref{fig:Xa},  the incorporation of the hierarchical structure among text-level covariates and auxiliary features can substantially improve the detection power.

In this paper, we present a hierarchical probabilistic black-box method which incorporates the test covariate hierarchy to control the FDR, named NeurT-FDR.  A
schematic diagram for NeurT-FDR is shown in Figure ~\ref{fig:illustration}.  NeurT-FDR builds on BB-FDR \cite{tansey18} and two-groups model \cite{efron01}. In two-groups model, a fraction of the test statistics is assumed to come from an unknown signal distribution, and the rest from a known null distribution. BB-FDR is a hierarchical probabilistic extension of the two-groups models to learn such mixing fraction by neural network from test-level covariates. Our proposal allows the mixing fraction to depend upon both the test-level and auxiliary covariates, and to estimate the form of this mixing fraction from the data.
 Our main contributions can be summarized as follows:
\begin{itemize}
\item We pioneer the use of auxiliary features associated with the  test-level covariates for multiple hypothesis testing problems.
\item We jointly learn the test-level covariates and its associated auxiliary features through a neural network to maximize discoveries. 
\item 
We developed a model that enables efficient optimization and gracefully handles high-dimensional hypothesis features. 
\end{itemize}

We provide extensive simulation studies on both synthetic and real datasets to demonstrate that our algorithm yields more discoveries while controlling the FDR compared to state-of-the-art methods.

\section{Related work}
The traditional methods for controlling the FDR, such as  Benjamini and Hochberg linear step-up procedure and Storey's q value, only use the $p$-values and impose the same threshold for all hypotheses. To increase the statistical power, many studies have been developed to take advantage of the test-level information \cite{scott15,ignatiadis16,xia17,zhang19}. The general formulations considered in these papers assume that each hypothesis has an associated feature vector (or called text-level covariates) related with the corresponding $p$-value. These features are supposed to capture some side information that might bear on the likelihood of a hypothesis to be significant, or on the power of $p$-value under the alternative, but the nature of this relationship is not fully known ahead of time and needs to be learned from the~data.

FDRreg \cite{scott15} is built on the two-groups model framework by taking into account the test-level covariates information to model the mixing fraction in a regression setting. IHW \cite{ignatiadis16} groups the hypotheses into a pre-specified number of bins according to their associated feature space and applies a constant threshold for each bin to maximize the discoveries. One major limitation of IHW on modern statistics is that binning the data into groups can be tremendously difficult if the feature space is of high-dimensional. NeuralFDR~\cite{xia17} addresses the limitation of IHW through the use of a neural network to parameterize the decision rule. This is a more general approach, and empirically it demonstrates that it works well on a multi-dimensional feature space. AdaFDR \cite{zhang19} is an extension of NeuralFDR which models the discovery threshold by a mixture model using the expectation-maximization algorithm. The mixture model is a combination of a generalized linear model and Gaussian mixtures and displays improved power in comparison with IHW and NeuralFDR. However, AdaFDR only work with multi-dimensional feature, as the number of parameters in AdaFDR grows linearly with respect to the covariate dimension. Thus, it is a 
substantial limitation for modern large-scale problems where a high-dimensional covariate setting is typical.

The recent work most relevant to ours is BB-FDR \cite{tansey18}. 
We think BB-FDR is the benchmark method for using a neural network to learn the true distributions of the test statistics from data in MHT. However, the existing model framework only deals with test-level covariates, while our method enables the learning from both test-level covariates and their associated auxiliary features.     

\section{Preliminaries \label{sec:prel}}
Consider the situation with $n$ independent hypotheses whereby each hypothesis $i$ produces a test statistics $z_i$ corresponding to the test outcome. Now, each hypothesis also has $k$ test-level covariates $\mathbf{X}_{i} = (X_{i1},...,X_{ik})' \in \mathcal{R}^k$ and $q$ auxiliary  covariates $\mathbf{X}^a_{i} = (X^a_{i1},...,X^a_{iq})' \in \mathcal{R}^q$  characterized by a tuple $(z_i, \mathbf{X}_i,\mathbf{X}^a_i,  h_i)$,  where $h_i \in \{0,1\}$ indicates if the $i$th hypothesis is null ($h_i=0$) or alternative ($h_i=1$) which depends on both $\mathbf{X}_{i}$ and $\mathbf{X}^a_{i}$. The test statistics $z_i$ is calculated using data different from $\mathbf{X}_{i}$ and $\mathbf{X}^a_{i}$.  The standard assumption is that under the null ($h_i = 0$), the distribution of the test statistic $z_i$ is from the null distribution, denoted by $f_0(z)$; otherwise $z_i$ follows an unknown alternative distribution, denoted by $f_1(z)$. The alternative hypotheses for  $h_i = 1$ are the \emph{true signals} that we would like to discover.

The general goal of multiple hypotheses testing is to claim a maximum number of discoveries based on the observations $\{(z_i, \mathbf{X}_i, \mathbf{X}^a_i)\}_{i=1}^n$ while controlling the false positives. 
The most popular quantities that conceptualize the false positives are the family-wise error rate (FWER) \cite{dunn61} and the false discovery rate (FDR) \cite{benjamini95}.
We specifically consider FDR in this paper.
FDR is the expected proportion of false discoveries, and one closely related quantity, the false discovery proportion (FDP), is the actual proportion of false discoveries.
We note that FDP is the actual realization of FDR. 
\subsection{False discovery rate control}
\label{subsec:background:fdr}
 For a given prediction $\hat{h}_i$, we say it is a true positive or a true discovery if $\hat{h}_i = 1 = h_i$ and a false positive or false discovery if $\hat{h}_i = 1 \neq h_i$. Let $\mathcal{D} = \{i : h_i = 1\}$ be the set of observations for which the treatment had an effect and $\hat{\mathcal{D}} = \{i : \hat{h}_i = 1\}$ be the set of predicted discoveries. We seek procedures that maximize the true positive rate (TPR) also known as \textit{power}, while controlling the false discovery rate -- the expected proportion of the predicted discoveries that are actually false positives.


\begin{definition} FDP and FDR

The false discovery proportion FDP and the false discovery rate FDR are defined as 
\begin{equation}
\label{eqn:fdp}
\text{FDR} \triangleq \coloneqq \mathbb{E}[\text{FDP}] \, , \quad \quad \text{FDP} \triangleq \frac{\#\{ i : i \in \hat{\mathcal{D}} \backslash \mathcal{D} \}}{\#\{ i : i \in \hat{\mathcal{D}} \}} \, .
\end{equation}
\end{definition}

In this paper, we aim to maximize ${\#\{ i : i \in \hat{\mathcal{D}} \}}$ while controlling $FDP\leq \alpha$ with high probability. 

\subsection{Two-groups Model}
\label{subsec:method:twogroups}
For the two-groups model, consider $n$ independent tests that have test statistics $z_1, \ldots, z_n$ as arising from a mixture model of two components
\begin{equation}
\label{eqn:z}
z_i \sim \lambda f_1(z_i) + (1-\lambda) f_0(z_i), \; \;i = 1,...,n. 
\end{equation}

where $\lambda \in (0,1)$ is the mixing proportion, and $f_0$ and $f_1$ describe the null ($h_i$ = 0) and the alternative ($h_i$ = 1) distributions of the test statistics, respectively. For each test statistics $z_i$, we are interested in reporting  

\begin{equation}
\label{eqn:posterior}
{w}_i = p(h_i = 1 | z_i) \\
=  \frac{\lambda f_1(z_i)}{\lambda f_1(z_i) + (1-\lambda) f_0(z_i)}.
\end{equation}

$w_i$ is interpreted as the posterior probability that $z_i$ is from the alternative distribution, while $1 - w_i$ is the local FDR. We reject $m$ hypotheses, where $0 \leq m \leq n$ is the largest possible index such that the expected FDP is below the FDR threshold. The global FDR of the test statistics $\mathbf{z} = (z_1, \ldots, z_n)$ can be estimated as
\begin{equation}
\label{eqn:step_down_procedure}
\begin{aligned}
& \text{FDR} \approx \frac{\sum_{i=1}^m (1-\hat{w}_i)}{m}.
\end{aligned}
\end{equation}
The limitation of the two-groups model is the assumption that all tests should be combined into a single analysis with a common mixing proportion $\lambda$ in (\ref{eqn:z}). Many studies  have demonstrated that the test-level covariates $X_i$ can affect the prior probability that $z_i$ is a signal. For example, in a genetic association study, each hypothesis test is interested in testing whether a mutation is correlated with the trait. For each test, the test-level covariates can include features both from mutation (e.g. location, epigenetic status etc) and the trait (e.g. gene expression value) \cite{scott15,tansey18,xia17,zhang19}. 


\section{Method\label{sec:method}} 
\subsection{NeurT-FDR model description}
NeurT-FDR extends the two-groups model \cite{efron08} and its hierarchical probabilistic extension \cite{tansey18} by learning a nonlinear mapping from the test-level covariates and their associated auxiliary features jointly to model the test-specific mixing proportion. More specifically, the model assumes a test-specific mixing proportion $\lambda_i$ which models the prior probability of the test statistics coming from the alternative (i.e. the probability of the test having an effect \textit{a priori}). Then, we place a Beta prior on each $\lambda_i$. In order to borrow information from both $\mathbf{X}_i$  and $\mathbf{X}^a_i$ when inferencing on $\lambda_i$, we propose to first learn a set of (pseudo) parameters, denoted by ($a'_i$, $b'_i$), of the Beta distribution with a deep neural network $G$ parameterized by $\theta_\phi$ from the test-level covariates $\mathbf{X}$. We then propose to adjust  
the learned parameters through a bivariate linear regression on the auxiliary features $\mathbf{X}^a$ to determine the parameters for Beta distribution, denoted by ($a_i$, $b_i$):
\begin{align}
(a'_i, b'_i) &= G_{\theta_\phi}(\mathbf{X}_{i})  \label{eqn:featureextraction1} \\
\left[\begin{matrix}
\,
  log_e(a'_i)\;
\\[2\jot]
\hfill log_e(b'_i)
\,
\end{matrix} \right] &\sim \mathcal {N}_2 \left\{ \left[
  \begin{matrix}
    \mu_a + \mathbf{X}^{a'}_i\boldsymbol{\delta}_a \\ \mu_b + \mathbf{X}^{a'}_i\boldsymbol{\delta}_b
  \end{matrix}
  \right], \left[
  \begin{matrix}
    \sigma_{aa} & \sigma_{ab}  \\
    \sigma_{ab} & \sigma_{bb}\\
  \end{matrix}
  \right] \right\} \label{eqn:featureextraction2}
\\
  (a_i, b_i) &= (\hat{a}'_i, \hat{b}'_i), \label{eqn:featureextraction3}
\end{align}
where $(a_i, b_i)$ is obtained as the fitted value from Eqn.~(\ref{eqn:featureextraction2}).

The complete hierarchical probabilistic framework for NeurT-FDR is then:

\begin{equation}
\label{eqn:generative_model}
\begin{aligned}
z_i &\sim h_i f_1(z_i) + (1-h_i) f_0(z_i) \\
h_i &\sim \text{Bernoulli}(\lambda_i) \\
\lambda_i &\sim \text{Beta}(a_i, b_i),
\end{aligned}
\end{equation}
where $(a_i, b_i)$ is obtained through Eqns.~(\ref{eqn:featureextraction1}) to ~(\ref{eqn:featureextraction3}). 
\begin{figure}[H]
\centering
\includegraphics[width=0.95\textwidth]{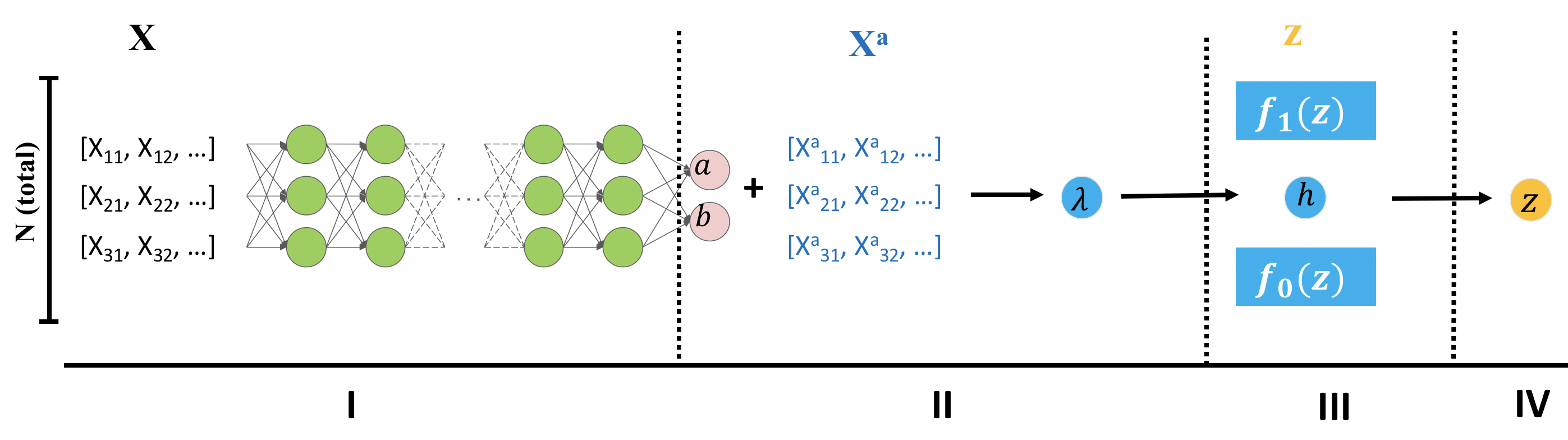}
\caption{The graphical demonstration for NeurT-FDR. {I}: The deep neural network learning from the test-level covariates ${\bf X}
$; {II}: The bivariate linear regression adjustment on the beta parameters learned from the test level covariates, a, b, with the auxiliary covariates ${\bf X}^a
$. {III}: Mixing the bernoulli prior $h$ with the estimated alternative distribution of $f_1(z)$ and $f_0(z)$ from the input $z$; {IV}: The learned statistic $z$ from data.}
\label{fig:model}
\end{figure}

\subsection{Estimation of the null and alternative distributions}

\begin{equation}
\label{eqn:null}
\begin{aligned}
f_0(z) & = \mathcal N(z|\mu,\sigma^2) \\
f_1(z) & = \int_\mathcal{R} \mathcal N(z|\mu + \tau, \sigma^2)\pi(\tau) d\tau
\end{aligned}
\end{equation}

Both $\mu$ and $\sigma^2$ are assumed to be known, but that can be relaxed by estimating an empirical null \cite{efron04}. To estimate $\hat{f}_1$, we use the predictive recursion \cite{newton02}. Eqn. (\ref{eqn:null}) is equivalent to formulation, \\
\begin{equation}
\begin{aligned}
z_i \sim \mathcal N(\mu + \tau, \sigma^2) \\
\Psi = \hat{\pi}_1(\tau) + h.\pi_0(\tau)
\end{aligned}
\end{equation}
Here, $\hat{\pi}_1(\tau) = h.\pi(\tau)$ is a sub-density corresponding to signals, and $\pi_0 = 1 - h$ is the mass at zero corresponding to nulls. 
Predictive recursion is a stochastic algorithm for estimating $\Psi$ from the test statistics input $z_1, \ldots, z_n$. Assume $\mu$ and $\sigma^2$ are fixed. The algorithm starts with an initial guess $\Psi^{[0]}$ and a sequence of weights $\gamma^{[i]} \in (0,1)$. For $i=1, \dots, n$, recursively compute the update: 

\begin{equation}
\label{eqn:recursive}
\begin{aligned}
m^{[i-1]}(z_i) = \int_\mathcal{R} \mathcal N(z|\mu + \tau, \sigma^2)\Psi^{[i-1]} (d\tau) \\
\Psi^{[i]}(d\tau) = (1-\gamma^{[i]})\Psi^{[i-1]}(d\tau) + \gamma^{[i]}.\left[\frac{N(z_i|\mu + \tau, \sigma^2)\Psi^{[i-1]}(d\tau) }{m^{[i-1]}(z_i)} \right]
\end{aligned}
\end{equation}

The final update, \\
\begin{equation}
\Psi^{[n]} = \hat{\pi}_1^{[n]}(\tau) + \pi_0^{[n]}\eta_0 = h^{[n]}.\pi^{[n]}(\tau) + (1-c^{[n]}).\eta_0
\end{equation}
This provides estimates for $h$ and the mixing density $\pi(\tau)$, the continuous component $\pi(\tau)$ is approximated on a discrete grid of points, and the integral in (\ref{eqn:recursive}) is computed by the trapezoid rule over this grid.
Figure \ref{fig:model} shows the NeurT-FDR graphical model. 

\subsection{Learning Inference}
By integrating out $h_i$ we get the objective function to maximize  $\theta_\phi$ as follows,
\begin{equation}
\label{eqn:bbfdr_data_likelihood}
\begin{aligned}
p_\theta(z_i) = \int_0^1 (\lambda_i f_1(z_i) + (1-\lambda_i) f_0(z_i)) \text{Beta}(\lambda_i \vert G_{{\theta}_\phi}(\mathbf{X}_i, \mathbf{X}^a_i)) d\lambda_i\,  .
\end{aligned}
\end{equation}



We fit the model in Eq. \eqref{eqn:featureextraction1}, \eqref{eqn:featureextraction2} and \eqref{eqn:generative_model} with stochastic gradient descent (SGD) on an $L_2$-regularized loss function,
\begin{equation}
\label{eqn:bbfdr_objective}
\begin{aligned}
& \underset{\theta \in {|\theta|}}{\text{minimize}}
& & 
-\sum_{i} \log
p_\theta(z_i) + \lambda_i {G_\theta(\mathbf{X}_i, \mathbf{X}^a_{i})}_F^2 \, ,
\end{aligned}
\end{equation}
where ${\cdot}_F$ is the Frobenius norm. In pilot studies, we found adding a small amount of $L_2$-regularization prevented over-fitting at virtually no cost 
of statistical power. For computational purposes, we approximate the integral in Eq \eqref{eqn:bbfdr_data_likelihood} by a fine-grained numerical grid. 

\subsection{FDR control}
Once the optimized parameters $\hat{\theta}_\phi$ are chosen, we calculate the posterior probability of each test statistic coming from the alternative,
\begin{flalign}
\label{eqn:bbfdr_posterior}
\hat{w}_i &= p_{\hat{\theta}}(h_i = 1 | z_i) \\\nonumber
&= \int_0^1 \frac{\lambda_i f_1(z_i) \text{Beta}(\lambda_i | G_{\hat{\theta}}(\mathbf{X}_i, \mathbf{X}^a_{i}))}{\lambda_i f_1(z_i) + (1-\lambda_i) f_0(z_i)} d\lambda_i \, .
\end{flalign}

To maximize the total number of discoveries, first, we sort the posteriors in descending order by the likelihood of the test statistics being drawn from the alternative. We then reject the first $m$ hypotheses, where $0 \leq m \leq n$ is the largest possible index such that the expected proportion of false discoveries is below the FDR threshold. Formally, this procedure solves the optimization problem,
\begin{equation}
\label{eqn:step_down_procedure_2}
\begin{aligned}
& \underset{m}{\text{maximize}}
& & 
m \\
& \text{subject to} & & \frac{\sum_{i=1}^m (1-\hat{w}_i)}{m} \leq \alpha \, ,
\end{aligned}
\end{equation}
for a given FDR threshold $\alpha$.

The neural network model $G$ uses the entire test-level feature vector $X_{i\cdot}$ of every test to predict the prior parameters and then get adjusted by the entire auxiliary feature vector $X_{i\cdot}^{a}$ over $\lambda_i$. The observations $z_i$ are then used to calculate the posterior probabilities $\hat{w}_i$. The selection procedure in \eqref{eqn:step_down_procedure_2} uses these posteriors to reject a maximum number of null hypotheses while conserving the FDR.
\begin{figure}[H]
\centering
\includegraphics[width=0.45\textwidth]{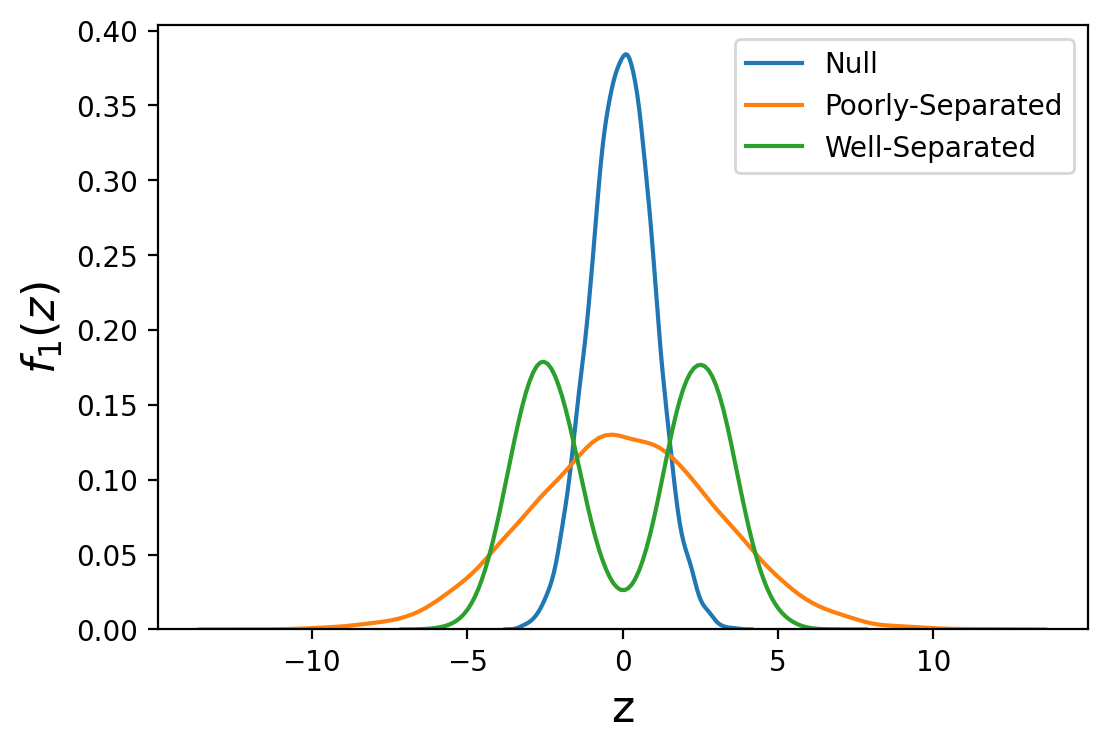}
\caption{The two alternative densities used in our simulation studies. The well-separated (WS) density has little overlap with the null, making for a stronger signal.}
\label{fig:alternative_densities}
\end{figure}

\section{Experiments\label{sec:emp}}
We evaluate our method using both simulated data and three real-world datasets.
For performance comparison, we consider BH \cite{benjamini95}, SBH\cite{storey04}, AdaFDR\cite{zhang19}, BB-FDR\cite{tansey18}, and two versions of our method, NeurT-FDRa, and NeurT-FDRb. We stack $\mathbf{X}$ and $\mathbf{X}^a$ together and put them into the $G_{\theta_\phi}$ for NeurT-FDRb. 

\subsection{Simulated data}
\label{subsec:benchmarks:setup}
{\bf Setup.} 
We consider three different ground truth models for $P(X)$ as used similarly in BB-FDR \cite{tansey18}, the joint distribution over the covariates, and $P(h=1 | X)$, the prior probability of coming from the alternative distribution given the covariates:
\begin{itemize}
    \item \textbf{Constant}: All covariates are sampled from independent and identically distributed normal; the prior is independent of the covariates, with $P(h_i=1 | X) = 0.5$.
    \item \textbf{Linear}: Covariates are sampled from a multivariate normal with identity covariance matrix; the prior is a linear function with IID standard normal coefficients for each covariate.
    \item \textbf{Nonlinear}: Covariates are sampled from a multivariate normal with identity covariance matrix; the prior is a nonlinear function with IID standard normal coefficients for each covariate.
\end{itemize}
For each of the three ground truth models, we consider two different alternative distributions:
\begin{itemize}
    \item \textbf{Well-Separated (WS)}: A 2-component Gaussian mixture model, $f_1(z) = 0.5\mathcal{N}(-2.5,1) + 0.5\mathcal{N}(2.5,1)$
    \item \textbf{Poorly-Separated (PS)}: A single normal with high overlap with the null, $f_1(z) = \mathcal{N}(0,3)$.
\end{itemize}
Figure~\ref{fig:alternative_densities} shows the densities used in the simulation study. To systematically quantify the FDP and power of all methods, we conducted extensive analyses 
of six combinations of synthetic data where we know the ground truth. Each experiment is repeated 10 times and 95\% confidence intervals are provided. We set the nominal FDR level $\alpha=[0.05, 0.10, 0.15,0.20]$. 
Each experiment, we generated 100 test-level covariates and 5 auxiliary features. For AdaFDR \cite{zhang19}, we use the auxiliary features since it cannot handle 
high dimensional test-level covariates.

\begin{figure*}
\centering
\vskip -0.3in
\subfigure[Constant (PS)]{\includegraphics[ width=0.32\textwidth]{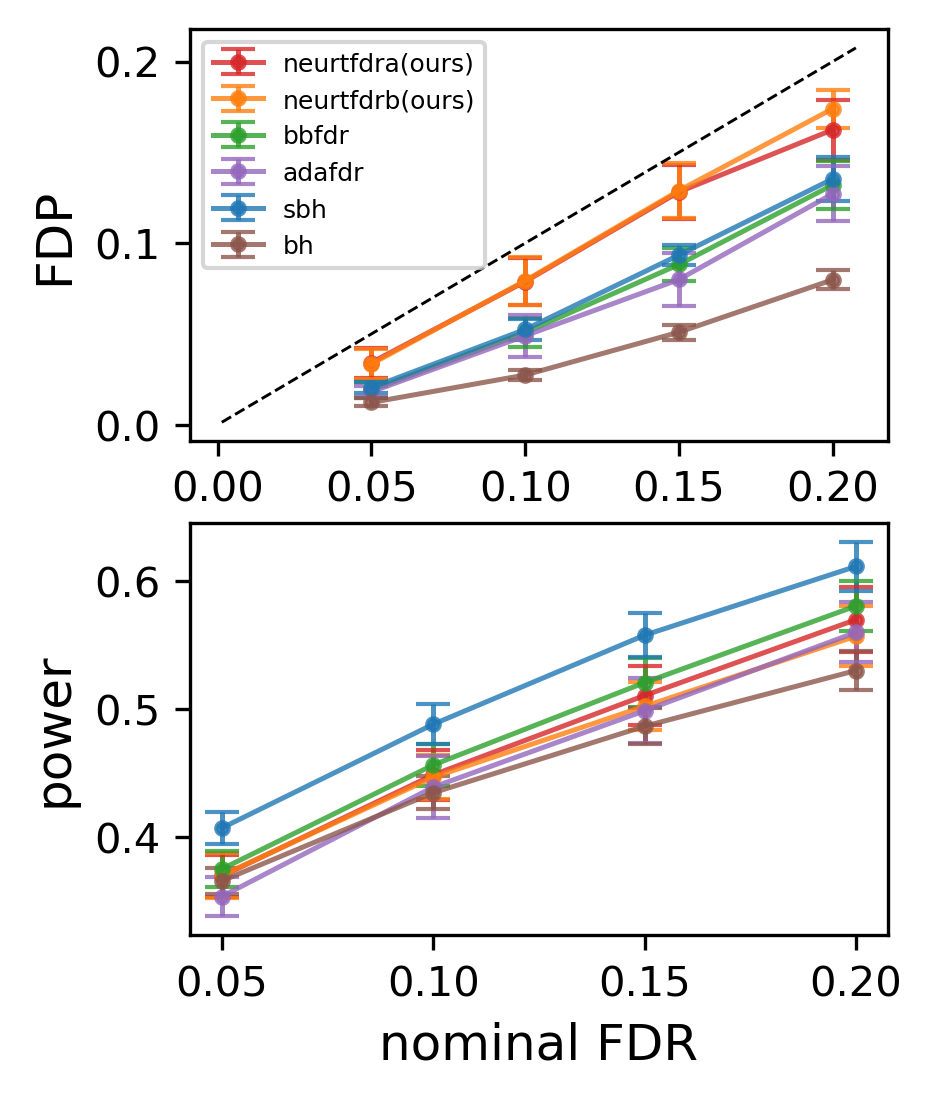}}
\subfigure[Linear (PS)]{\includegraphics[ width=0.32\textwidth]{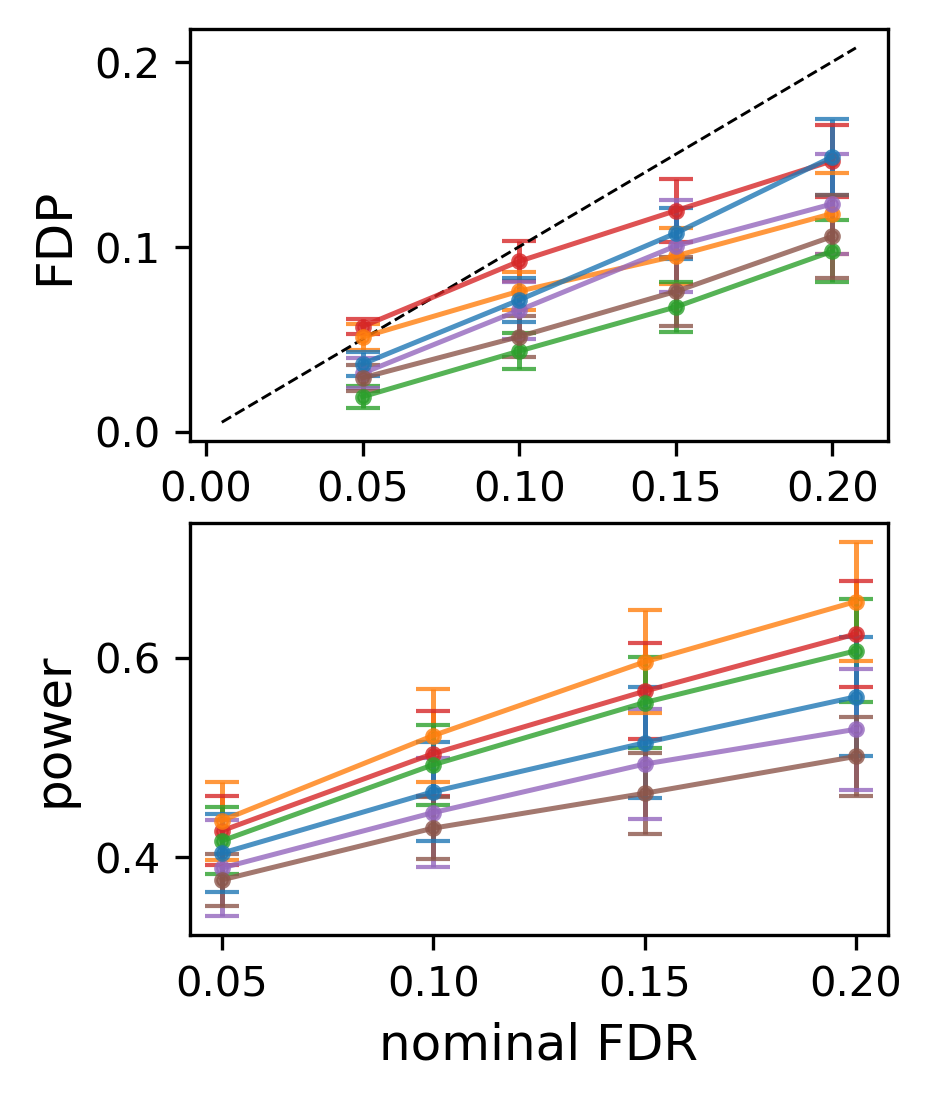}}
\subfigure[Nonlinear (PS)]{\includegraphics[width=0.32\textwidth]{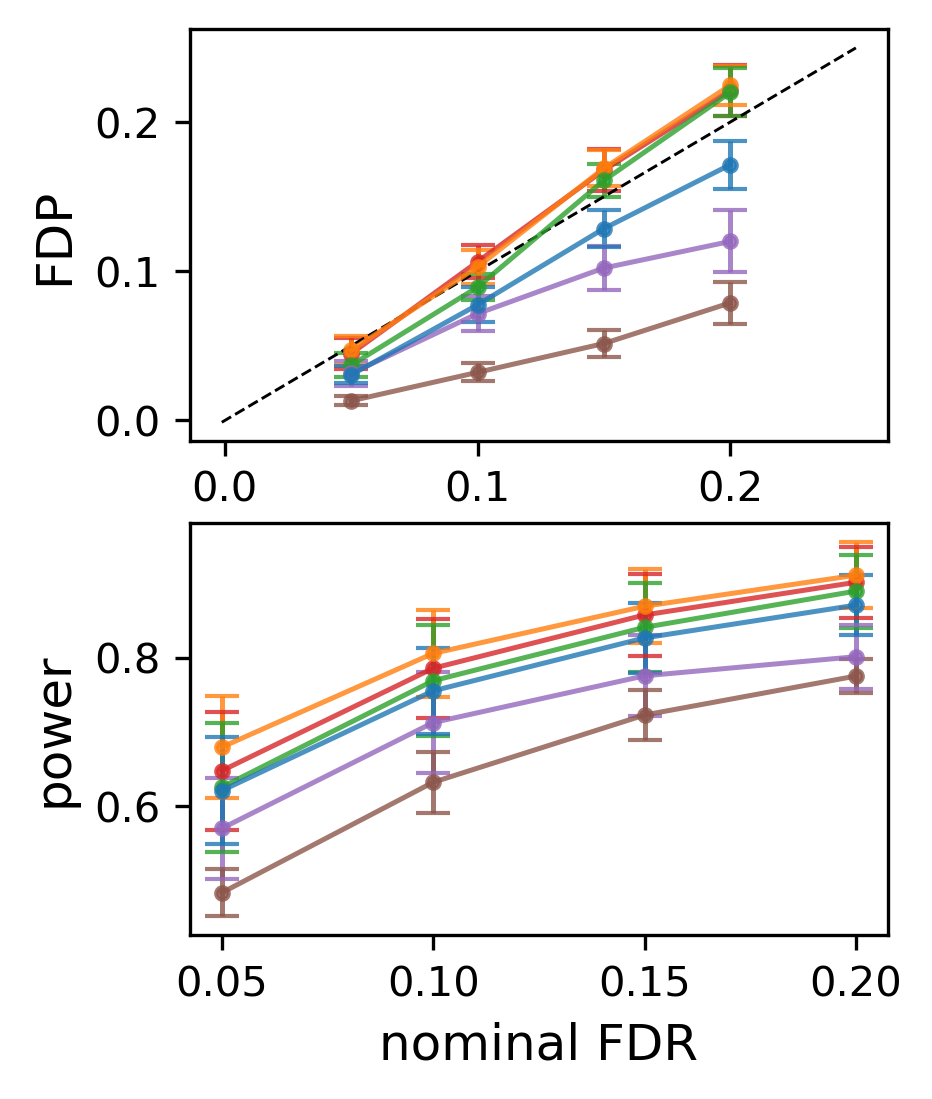}}
\subfigure[Constant (WS)]{\includegraphics[width=0.32\textwidth]{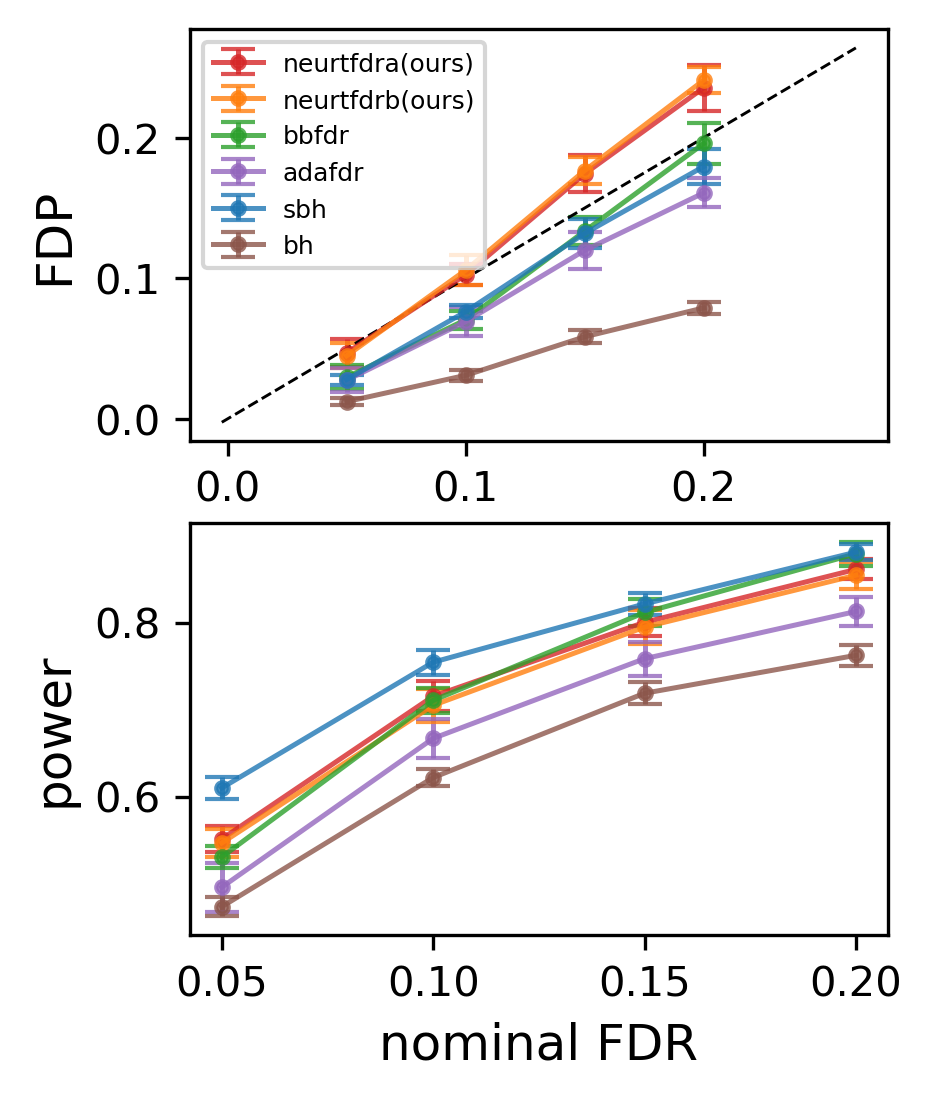}}
\subfigure[Linear (WS)]{\includegraphics[ width=0.32\textwidth]{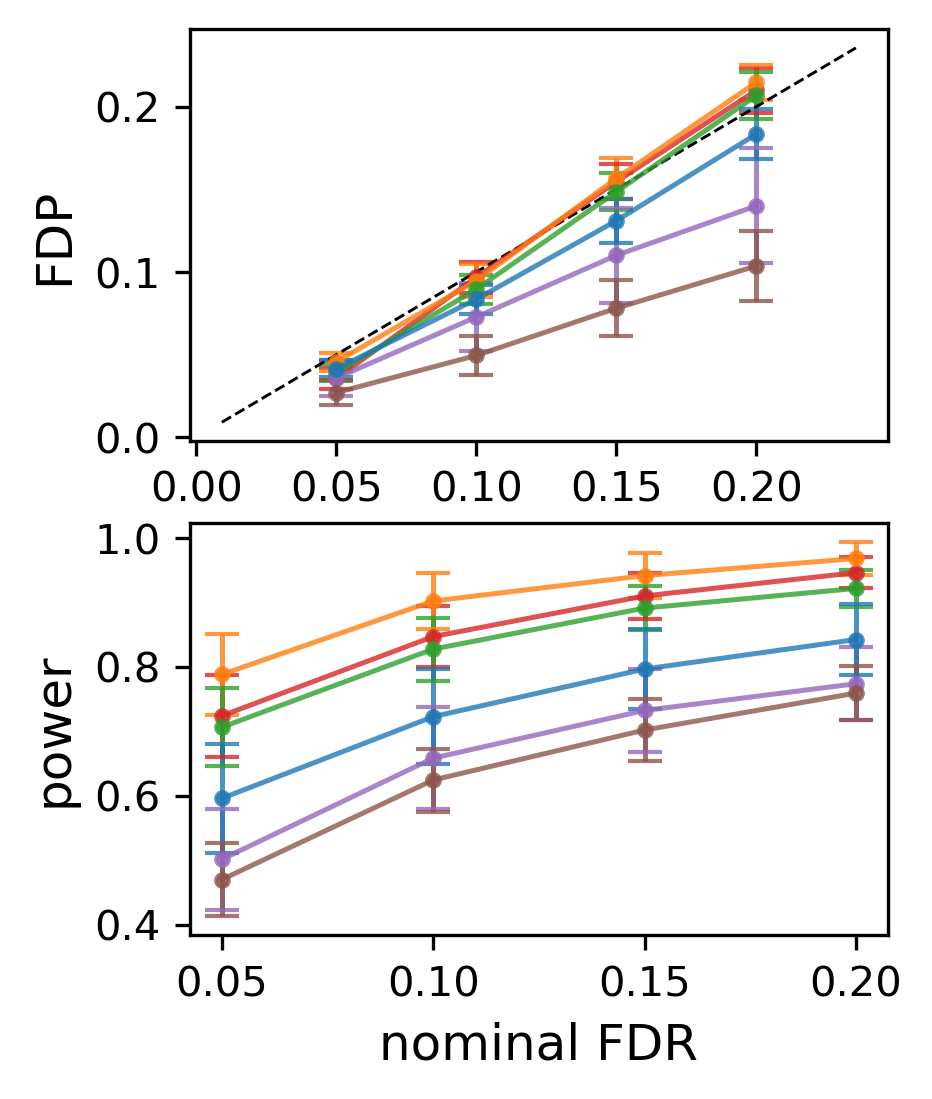}}
\subfigure[Nonlinear (WS)]{\includegraphics[ width=0.32\textwidth]{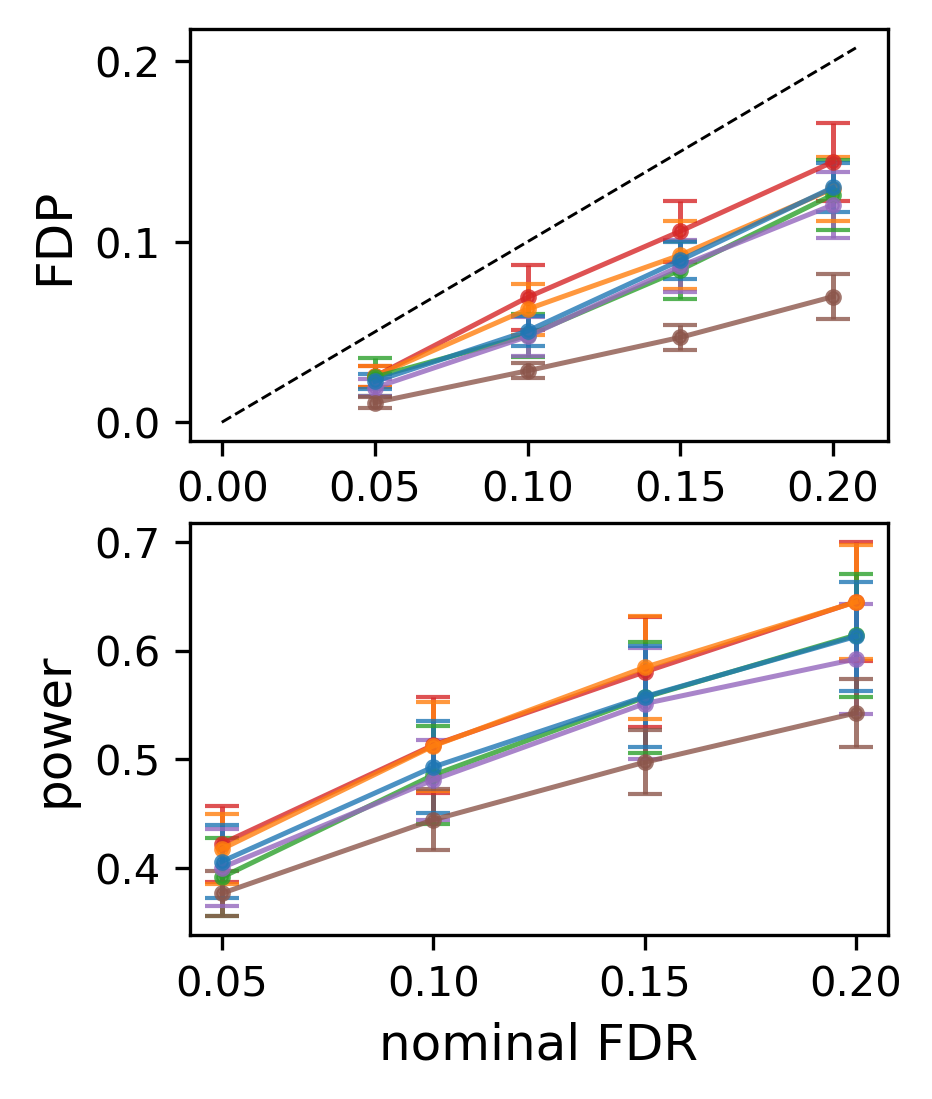}}
\caption{\label{fig:simulation} Hypothesis testing results on the synthetic datasets averaged over 10 trials at sample size $n=1000$ on the two different alternative distributions (Poorly-separated (PS): first two rows and Well-Separated (WS): last two rows ). In general, NeurT-FDRa and NeurT-FDRb have higher power since they model the alternative under linear and nonlinear relationships with the covariates. Constant, Linear, Nonlinear represents the three ground truth model for $P(X)$ and PS, WS indicates 
Poorly-Separated and Well-Separated
alternative distributions.}

\end{figure*}
\begin{figure}
\centering
\vskip -0.3in
\subfigure[Nonlinear (PS)]{\includegraphics[width=0.40\textwidth]{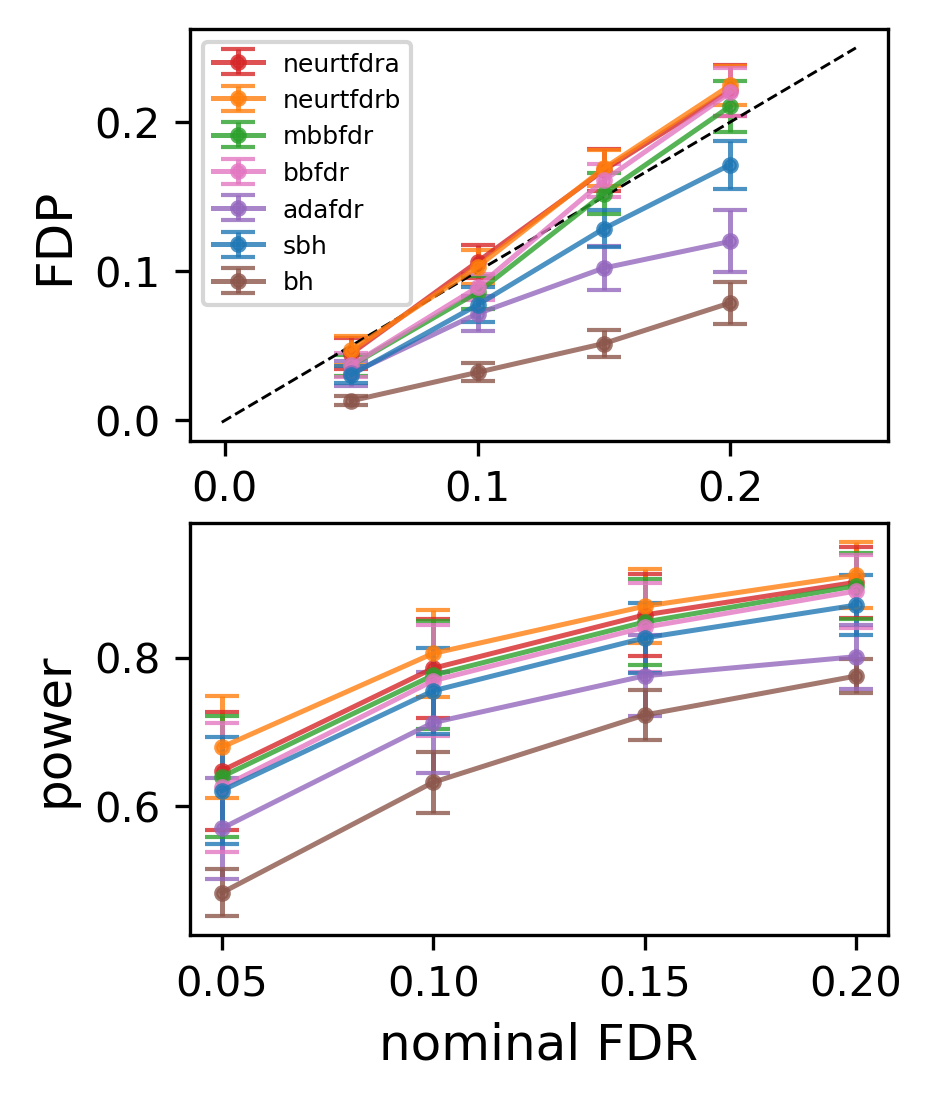}}
\subfigure[Nonlinear (WS)]{\includegraphics[ width=0.40\textwidth]{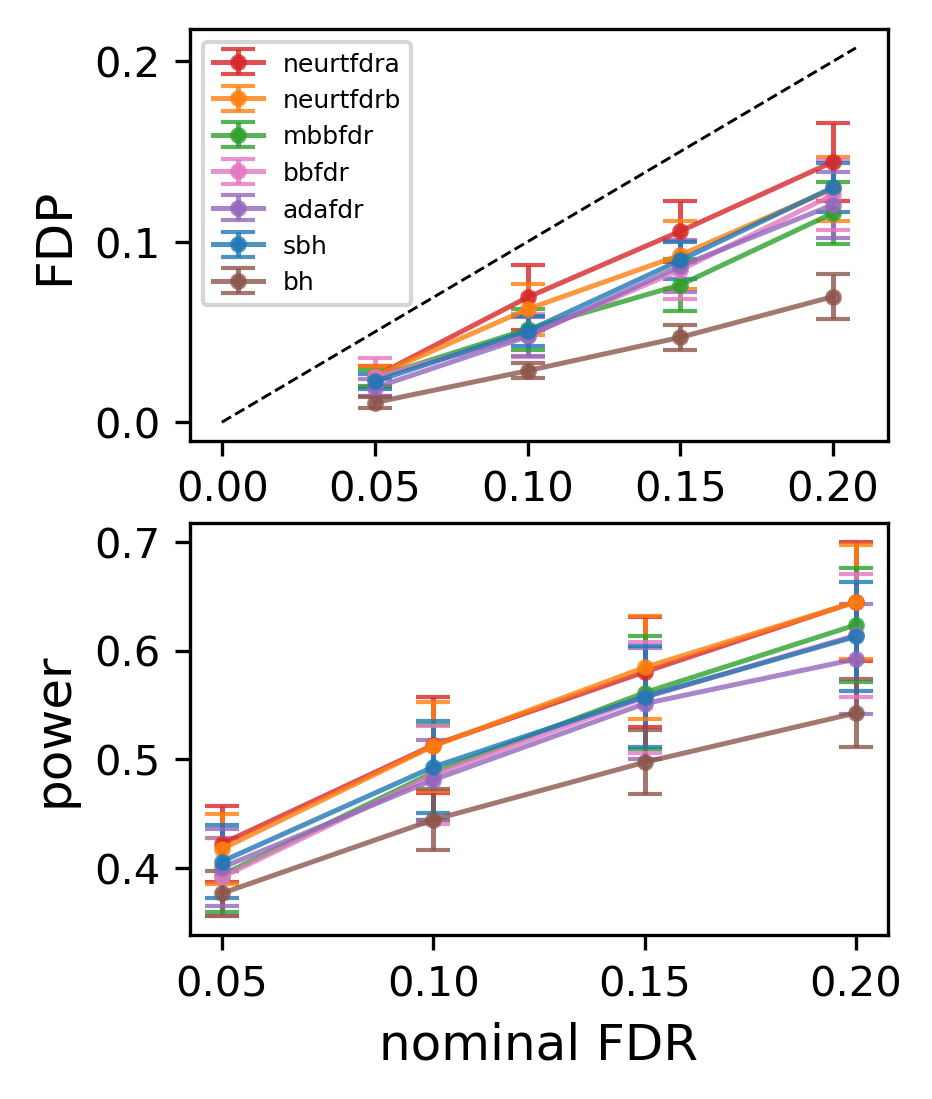}}
\caption{\label{fig:s1} Hypothesis testing results on the synthetic datasets averaged over 10 trials at sample size $n=1000$ on the nonlinear scenario for two different alternative distributions (Poorly-separated (PS) and Well-Separated (WS)). }
\end{figure}
{\bf Evaluation metrics.} We report the FDP as discussed in section \ref{subsec:background:fdr} and Power. 

{\bf Results.} Under linear and nonlinear relationship scenarios, from Figure \ref{fig:simulation}, NeurT-FDRa and NeurT-FDRb outperform other methods which achieve significantly higher power and better FDR control, BB-FDR performs slightly worse than NeurT-FDRa regarding power. Under a constant scenario, SBH has the highest power for Poorly-Seperated and Well-Separated alternative distributions while NeurT-FDRa and NeurT-FDRb has better FDR control. This 
result reflects that BB-FDR, AdaFDR, and our method are all developed based on test-level covariates, and under no test side information SBH performs best. Our method achieves the best FDR control indicating that the model's power 
captures related auxiliary feature information. AdaFDR performs not very well under all conditions which means AdaFDR is limited for low dimension applications. Furthermore, we stacked the test-level covariates and auxiliary features together and applied the BB-FDR model. We named this method as MBB-FDR.  Figure \ref{fig:s1} shows that MBB-FDR (green color) performs better than BB-FDR but not as well as
NeurT-FDRa and NeurT-FDRb, which indicates that the auxiliary features contributed the test signal. As a result, it should be modeled differently from the test-level covariates.

\subsection{Real data}
 We consider three real-world datasets: Two drug screening studies (Lapatinib and Nutlin-3) from the Genomics of Drug Sensitivity in Cancer (GDSC) \cite{yang12} used in BB-FDR, which aims to investigate how cancer cell lines respond to different cancer therapeutics; We consider Airway\cite{himes14} as the RNA-Seq dataset which is widely used in other multiple testing studies like IHW, AdaFDR, and NeuralFDR. This study aims to identify glucocorticoid responsive genes which regulate cytokine function in airway smooth muscle cells; finally, we experiment with two Snap datasets in which one contains
$n$=10,740 public Snaps from Female, North America, US, 20200902; and the other contains $n$=62,406 public Snaps on labor day, North America, US, 20200907. Essentially, these datasets contain visual tag annotations and content consumption metrics related to user interactions in the corresponding social media platform. In this case, our method can be applied to understand how users engage and interact with Snap contents for improving user retention and personalized content recommendation. \\ 

\begin{figure}[th]
\centering
\subfigure[\small BH\cite{benjamini95} on Lapatinib  \newline \text{  }(151 discoveries)]{\includegraphics[width=0.35\textwidth]{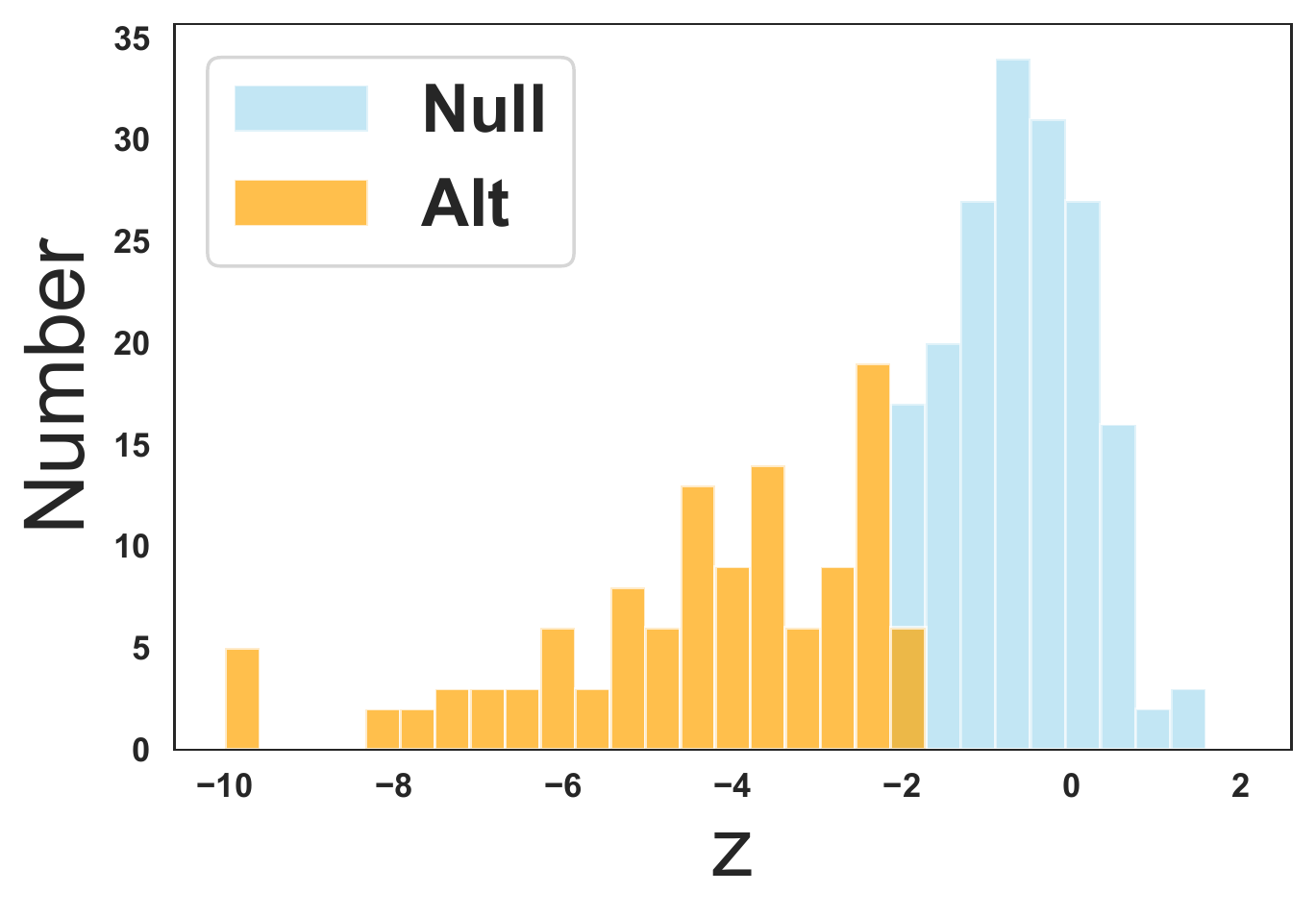}}
\subfigure[\small BH\cite{benjamini95} on Nutlin-3 \newline  \text{  } (117 discoveries)]{\includegraphics[width=0.35\textwidth]{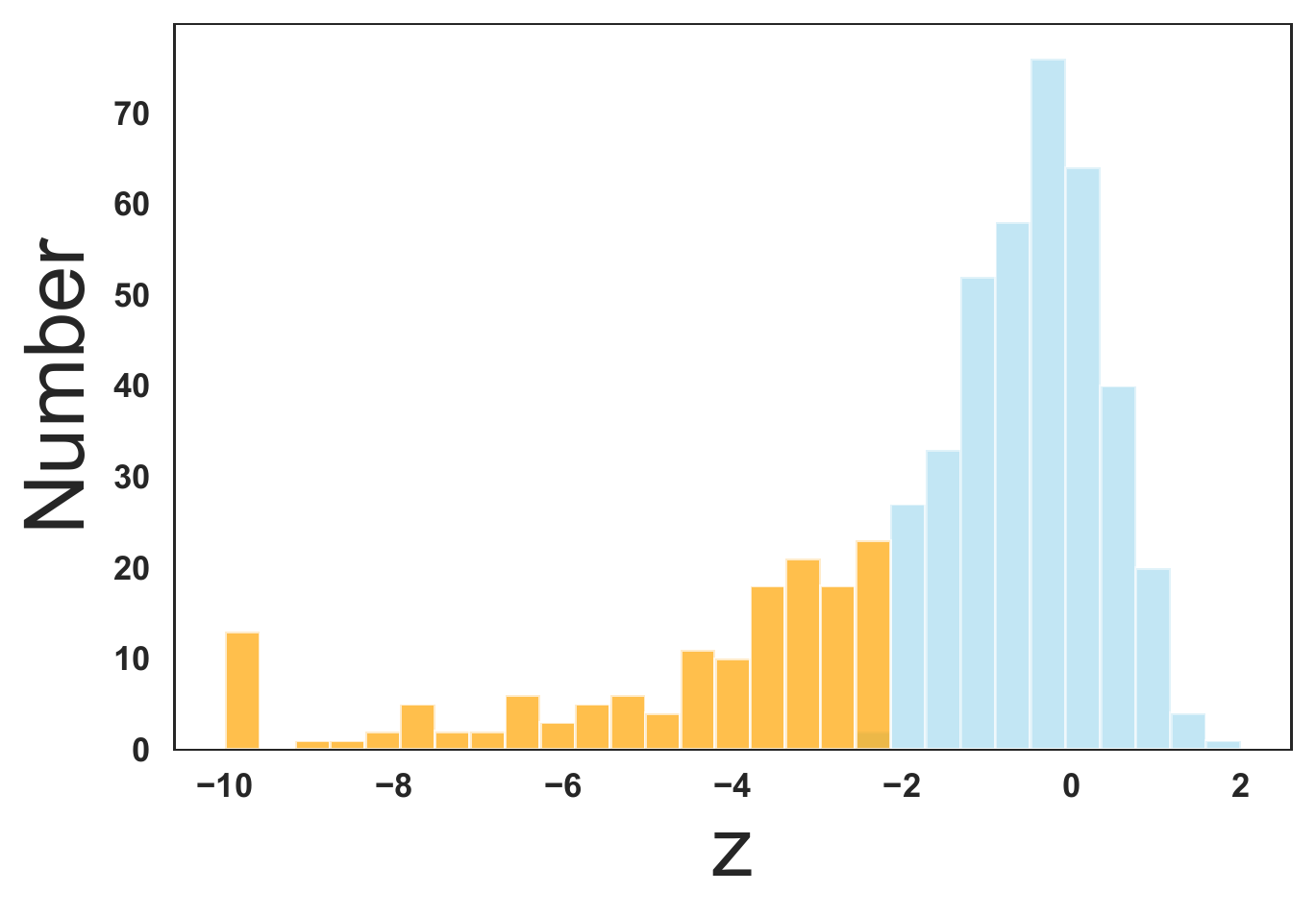}}
\subfigure[\small NeurT-FDRa on Lapatinib \newline \text{  }(187 discoveries)]{\includegraphics[width=0.35\textwidth]{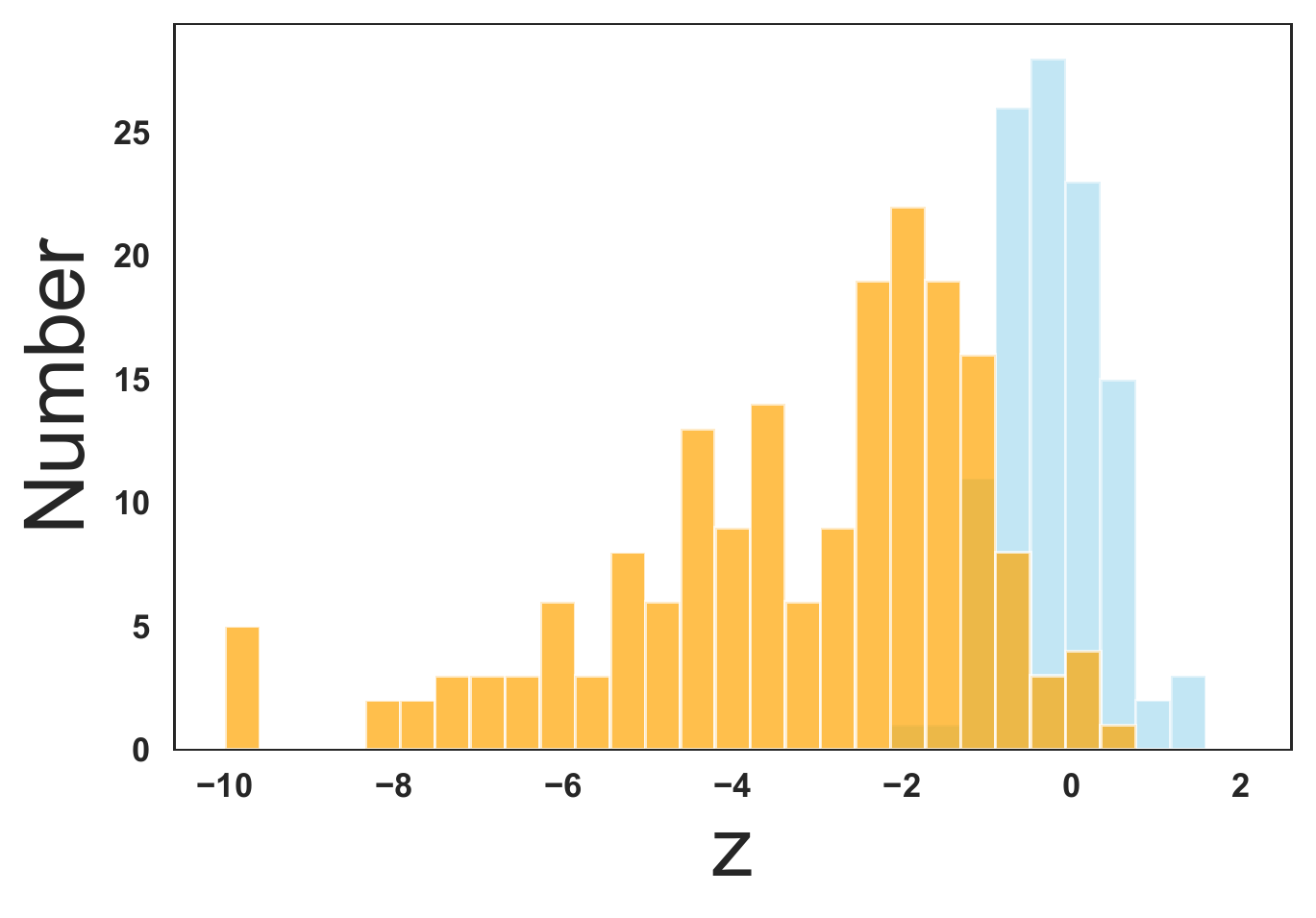}}
\subfigure[\small NeurT-FDRa on Nutlin-3 \newline  \text{  } (215 discoveries)]{\includegraphics[width=0.35\textwidth]{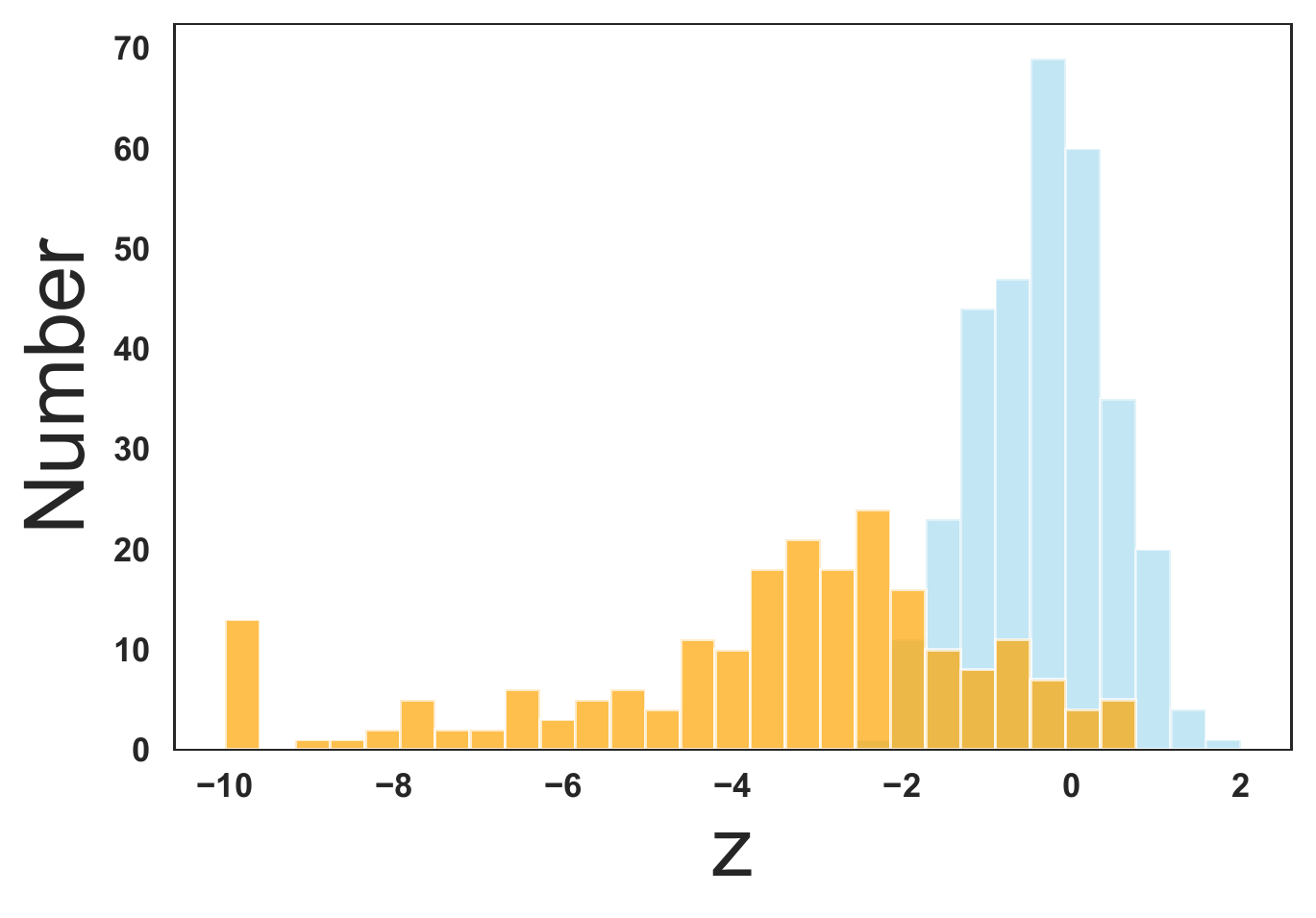}}
\caption[Z-scores for the cancer drug case study]{\label{fig:cancer_discoveries} Discoveries found by NeurT-FDRa on the two drugs, compared to the discoveries found by a naive BH\cite{benjamini95} approach. Blue and orange represents the null and alternative discoveries respectively.}
\end{figure}

{\bf Cancer drug screening data.}
 One goal of this analysis is to address the question of whether a given cell line responded to the drug treatment. Thus, this is a classical multiple testing problem that we need a hypothesis test for each cell line, where the null hypothesis is that the drug had no effect. We use the data preprocessed by \cite{tansey18} which contains genomic features and the z-score relative to mean control values for each cell line. We treat the genomic features as the test-level covariates and extract the rank of the z-score as the auxiliary feature. For AdaFDR, we use the auxiliary features for the covariate input. Table \ref{tab:realdata} shows for both drugs NeurT-FDRa and NeurT-FDRb achieve the largest power compared to other methods and Figure \ref{fig:cancer_discoveries} shows that the test-level covariates and auxiliary features provide enough prior information that even some outcomes with a z-score above zero are still found to be significant in NeurT-FDRa.
 
\begin{table}[th]
\centering
\caption{Real data: \# of discoveries at FDR = 0.1. Best two performers per dataset are highlighted in bold.}
\label{tab:realdata}
\begin{tabular}{l|l|l|l|l}
          & Lapatinib \cite{tansey18} & Nutlin-3 \cite{tansey18}  & Airway \cite{zhang19}   & Visual Tags\\
\hline
BH \cite{benjamini95}             & 117   & 151  & 4,079 & 312\\
SBH \cite{storey04}         &   131(+11.9\%)   & 159(+5.3\%)  & 4,079 & 312\\
AdaFDR \cite{zhang19}       & 137(+9.7\%)  &   161(+37.6\%)    & {\bf 6,050(+48.3\%)} & -  \\
BB-FDR \cite{tansey18}  & 181(+54.7\%)  &  210(+39.1\%)  & 5,791(+41.9\%) & 385(+23.4\%) \\
NeurT-FDRa (\textbf{ours})  &{\bf 187(+59.8\%)}  & {\bf 215(+42.3\%)}  &5,859(+43.6\%) & {\bf 389(+24.7\%)}\\
NeurT-FDRb (\textbf{ours})  & {\bf 212(+81.2\%)}  & {\bf 260(+72.2\%)}  &{\bf 8,820(+116\%)} & {\bf 593(+91.9\%)}\\
\end{tabular}
\end{table}


{\bf RNA-Seq data.}
 The original dataset contains a p-value and a log count for each gene (n=33,469), we consider the log count for each gene as the test-level covariate and the rank for the p-value as the auxiliary feature. As the result shown in Table \ref{tab:realdata}, where BB-FDR, and NeurT-FDRa have similar number of discoveries, AdaFDR performs slightly better and NeurT-FDRb provides 50\% more discoveries than all of them. All covariate-related methods make significantly more discoveries than the non covariate-related methods. NeurT-FDRb achieves 116\% more discoveries compare to BH even when the dataset contains only one test-level covariate.   

{\bf Snap Visual Tags data.} 
 Each Snap has a visual tags vector coming from computer vision models with its corresponding content consumption metrics like how much time the particular user group spent on this Snap, number of shares, number of views, and others. So, we investigate which Snaps has the top engagements when they are compared to the normal behavior in one particular user cohort (i.e., age group and gender specification). We consider the visual tags as the test-level covariate, and the associated 16 content consumption metrics as the auxiliary features. We used z-score as the ratio between Snap view time ratio and the number of view records to the mean values for each Snap. From our results in Table \ref{tab:realdata}, NeurT-FDRa and NeurT-FDRb provides significantly more discoveries than other methods, and here AdaFDR failed because it only can handle very low dimensions of covariates. AdaFDR worked in the cancer drug screening and RNA-seq data analysis when we used the rank of the test statistics as feature input. However, here we have 16 associated content consumption metrics which is a big advantage to our method since it is capable of handling both high-dimensional test and auxiliary level features' hypothesis test.

\begin{figure}[th]
\centering
\subfigure[]{\includegraphics[width=0.36\textwidth]{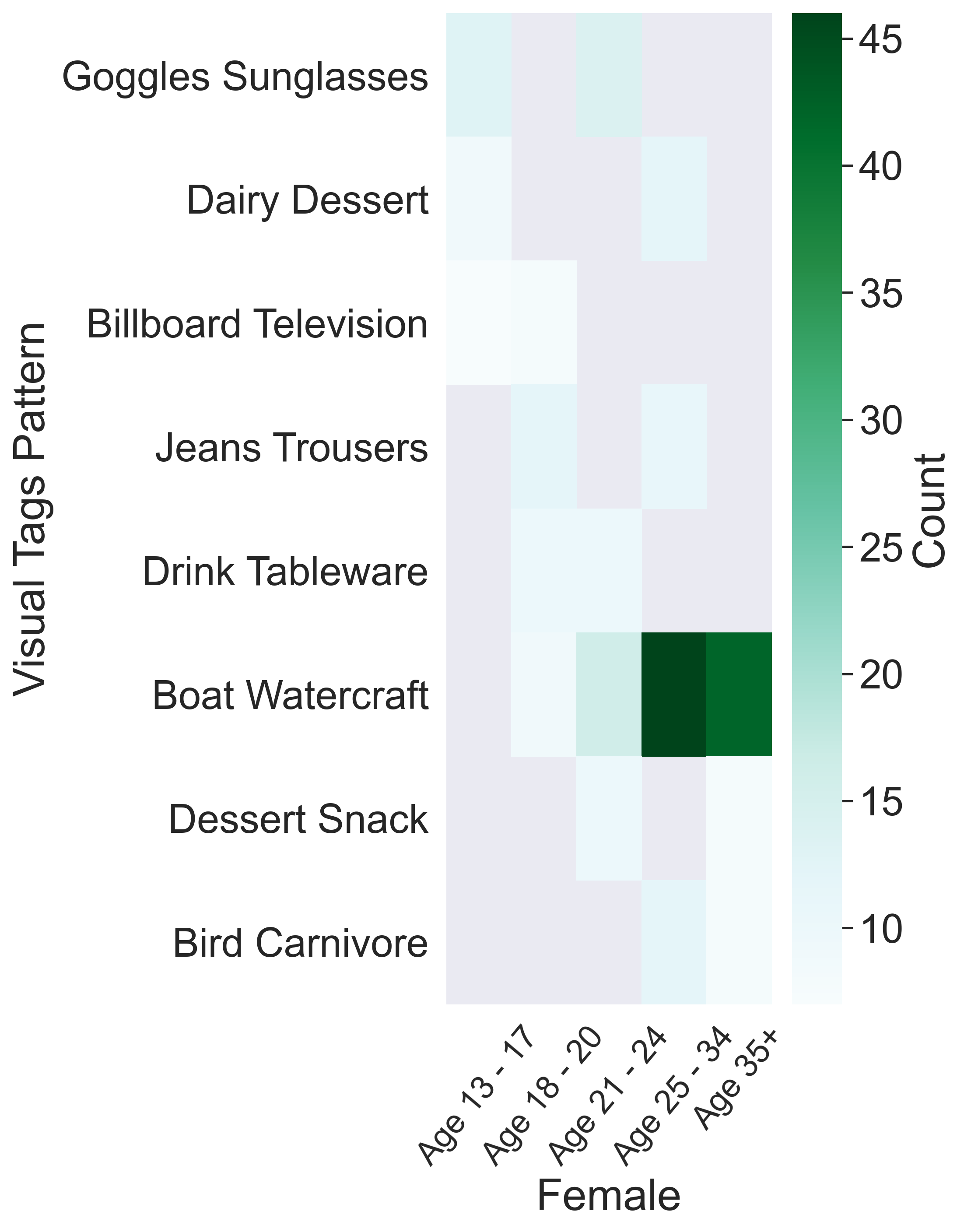}}
\subfigure[]{\includegraphics[width=0.325\textwidth]{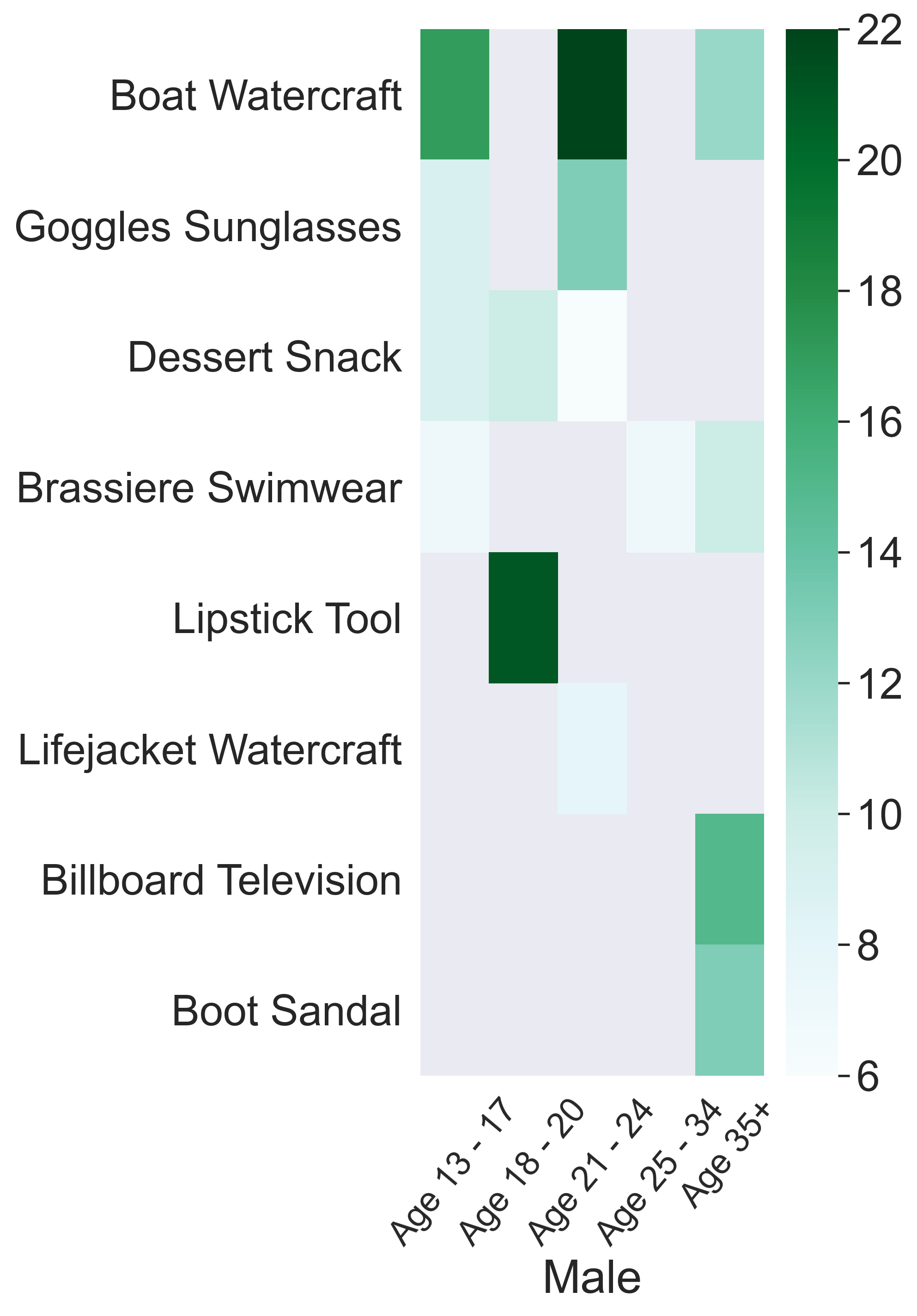}}

\caption{\label{fig:visualtags} Top visual tags pattern. {(a)}: Female users. {(b)}: Male users.}
\end{figure}

Another interesting application of our model is to extract the patterns with popular visual tags. By sorting the posterior probability in descending order, we gain the list of the most engagement Snaps for a particular group of user at a particular time window. Similarly, BB-FDR also can do this, however, our model is more focused on the content consumption metrics not just only on the visual tags which is more related with the user engagement behavior. In this analysis, we experiment the Snap dataset in a festivity date with NeurT-FDRa which we can have the popularity list of all the Snaps and determine the most frequent visual tags pattern by our internal exhaustive search based data mining algorithm. This finding is very important to understand the user engagement and to provide content recommendation. Figure \ref{fig:visualtags} shows the interesting findings based on 2 visual tags for each pattern, indicating engagement preferences arising from gender and age.

For example, in Figure \ref{fig:visualtags} (b), we observe that male viewers are engaging with (boat, watercraft) and (googles, sunglasses). These patterns can relate to a vacation day at the beach or in a swimming pool. These results also make sense for parents date (our festivity date selected), where parents can take their children to enjoy family time at beaches or swimming pools.



\section{Discussion} To the best of our knowledge, NeurT-FDR is the first model which considers the feature hierarchy to maximize the discoveries in multiple hypothesis testing. We showed that NeurT-FDR 
controls FDR and makes more discoveries on synthetic and real datasets with high-dimensional features. We believe the main difference between NeurT-FDR and other methods which involve test-level information (e.g., BB-FDR, AdaFDR, and NeuralFDR) is that we do not only propose to consider the feature hierarchy in the model, we apply two-stage learning strategy to first learn the test-level covariates from the neural network and then get the learned model parameters adjusted by linear regression on the auxiliary covariates. 
The neural network embedding architecture for the test-level covariates and the linear regression model for learning the auxiliary covariates together enable NeurT-FDR to address modern high-dimensional problems. 

We believe NeurT-FDR will contribute to this field as a benchmark work for further investigation and have a wide application in genetics, neuroimaging, online advertising, finance, and social media.

\bibliographystyle{plain}
\bibliography{ref}
\end{document}